# Neuromorphic Processing & Sensing:
# Evolutionary Progression of AI to Spiking

Philippe Reiter, Geet Rose Jose, Spyridon Bizmpikis, Ionela-Ancuța Cîrjilă

*Abstract*— **The increasing rise in machine learning and deep learning applications is requiring ever more computational resources to successfully meet the growing demands of an always-connected, automated world. Neuromorphic technologies based on Spiking Neural Network algorithms hold the promise to implement advanced artificial intelligence using a fraction of the computations and power requirements by modeling the functioning, and spiking, of the human brain. With the proliferation of tools and platforms aiding data scientists and machine learning engineers to develop the latest innovations in artificial and deep neural networks, a transition to a new paradigm will require building from the current well-established foundations. This paper explains the theoretical workings of neuromorphic technologies based on spikes, and overviews the state-of-art in hardware processors, software platforms and neuromorphic sensing devices. A progression path is paved for current machine learning specialists to update their skillset, as well as classification or predictive models from the current generation of deep neural networks to SNNs. This can be achieved by leveraging existing, specialised hardware in the form of SpiNNaker and the Nengo migration toolkit. First-hand, experimental results of converting a VGG-16 neural network to an SNN are shared. A forward gaze into industrial, medical and commercial applications that can readily benefit from SNNs wraps up this investigation into the neuromorphic computing future.**

## I. Introduction

As early as the middle of the last century, researchers have theorized computational models for mimicking the operations of the human brain. These were the beginnings of neuromorphic or brain-like computing.

The pivotal turning point occurred in 1958, when, working out of the Cornell Aeronautical Laboratory, Frank Rosenblatt invented what is considered one of the first-known learning machines, the Perceptron. Born of this early binary classification neural network (NN), advances in the field of machine learning have ebbed and flowed through the years.

The last couple of decades have seen a more rapid rise in machine learning and deep learning (MLDL) developments. This accelerated pace of innovation was spurred on by the seminal works of LeCun, Hinton and others in the 1990s on convolutional neural networks, or CNNs. Since then, numerous, more advanced learning models have been developed, and neural networks have become integral to industry, medicine, academia and commercial devices. From fully autonomous vehicles to rapid facial recognition entering popular culture to enumerable innovations across almost all domains, it is not an exaggeration to claim that CNNs and their progeny have changed both the technological and physical worlds.

CNNs can be exceedingly computationally intensive, however, as will be explained, often limiting their use to high-performance computers and large data centres. As MLDL applications span a vast range of domains, they will be increasingly required in lower-performing mobile, edge and Internet-of-Things (IoT) devices. Spiking neural networks present a biologically-plausible evolution to CNNs, and their reduced-power and lower-latency characteristics enable their deployment into even minimally-performing computing systems. The necessary knowledge foundation, and a facilitated migration path for MLDL practitioners to transfer their expertise to spiking models are presented in this paper.

## II. Evolution To Spiking Neural Networks

The rate of inventions and innovations is only accelerating. As computational capabilities and MLDL discoveries increase, so does the ability for neuroscience to better understand the brain. A positive feedback loop is formed with ground-breaking neuroscience research accelerating MLDL ever-forward, and vice-versa.

### A. Inspired by the Brain

Complex deep neural networks (DNNs), which include CNNs, are increasingly effective at the exponential number of tasks to which they are being applied. This has the side effect of also requiring a similar staggering increase in heavily energy- and space-consuming computing systems to power these DNN algorithms.

As convolutional networks were inspired by the understanding of the brain's processing of information through a number of layers, each extracting increasingly finer details, the future of MLDL will again rest on leading-edge neuroscience discoveries.

Comparing modern computers to the human brain, the latter has evolved to be an exceedingly efficient, self-adapting and low-power sensing and learning system. Processing in the brain is highly distributed, parallel and sparse, with learning occurring at each synapse and with each neuron being connected to tens-of-thousands of others. Contrast this to the ten-to-one or less fanout in a central processing unit (CPU) [1], which is often referred to the "brain" of a computer and which centrally controls learning. Moreover, the brain captures information in an event-based manner, with the processing and storage of information being time-encoded. This temporal input and computing stimuli are known as "spikes," and they are fundamental to Spiking Neural Networks (SNN) and crucial for developments in neuromorphic sensing and processing technologies.

In addition to new algorithms, inventive new hardware architectures are required to both capture real-world physical events as spikes, as well as drive these spikes through innovative architectures that excel at SNN computations. From analog to mixed-signal and purely digital implementations, the neuromorphic processors powering the SNN revolution are described in the *State of Neuromorphic Hardware* section, below. A special focus is given to the SpiNNaker implementation, as it directly arises from using an ingenious repurposing of existing technology. This theme of using current, widely-available technology as a springboard to future neuromorphic developments will resonate throughout this paper.

### B. Path to SNN Adoption

With thousands upon thousands of data scientists and machine learning engineers proficient with today's MLDL tools and DNN algorithms, a path has to be paved to transition this expertise to also leverage low-power, low-latency and highly parallel SNN models, where appropriate. A novel suite of new software is becoming available and is being developed to address the need for a clear roadmap from where MLDL deployments are occurring now to the desired state of neuromorphic devices in the future.

From professional development platforms to community toolkits, and from simulators and debuggers to emulators, a varied spectrum of SNN software is available to facilitate the transition from conventional CNN and general DNN methodologies to SNN models. The panoply of neuromorphic software options is overviewed in the *State of Neuromorphic Computing* section later in this paper. This is followed by first-hand, real-world experimental results of rapidly converting a conventional CNN, VGG-16, to an SNN using the Nengo toolkit from Applied Brain Research.

Finally, the applications of neuromorphic technology and devices to today's world and that of the future are covered in the wrap-up to this paper. There are presently exciting uses of neuromorphic hardware and SNN implementations in autonomous systems, robotics and drones, vision and recognition sensors, and edge devices. With artificial intelligence (AI) and MLDL becoming an intrinsic component of, not only the computing field, but the very fabric of the modern world's intelligence, security and social infrastructure, the shortlist of future applications presented comprises only a sampling of the advancements to be realised in the coming years.

To establish the groundwork for the rest of this paper's neuromorphic exploration, the following section overviews the neuroscientific and mathematical underpinnings that led to the development of SNNs.

### III. FUNDAMENTALS OF SNNs

As mentioned, the lightning advances in CNN research and machine learning applications have come at the expense of massive, expensive and power-hungry cloud server operations.

CNN algorithms require the highest in computational performance to quickly execute. This kind of performance is not available on the devices that most leverage the benefits of MLDL networks – distributed mobile and edge smart devices.

Revisiting the biological functioning of the brain, it was determined that neurons quantise impulses as either "ON" or "OFF" – not as the smooth, highly precise, fractional quantisation found in CNN weights.

The modeling of these ON/OFF biological neural spikes has led to numerous research institutions and large technology companies around the world investing in Spiking Neural Network (SNN) algorithms and the burgeoning field of neuromorphic computing – that is, "brain-like" computing.

The spiking properties of neuromorphic software and hardware promise significant reductions in power consumption and processing delays, compared to current CNNs and DNNs. The human brain consumes on average 20W, and it is expected that neuromorphic developments mirroring the brain's operation can achieve comparable levels of low-power performance. Similar to the brain, SNNs are massively parallel, which further benefits rapid computations, especially that of NN calculations.

Methods of generating spikes and converting existing CNNs and DNNs are specific to the approach applied for building an SNN. The most stable techniques are explained in the following section.

### A. Spike Theory Overview

The most readily implementable SNNs today are those based on converted DNNs. To obtain an SNN, [2] suggests to modify DNNs according to the neuromorphic paradigm by either using binary activation functions instead of the typical ones; training the DNNs with backpropagation, then converting the analog neurons into spiking ones; or, training the DNNs with pre-defined constraints that model the properties of the spiking neurons.

Other approaches to design and implement SNNs include a supervised, direct training of the SNNs with variations on error backpropagation; or, an unsupervised spike-timing-dependent plasticity (STDP) algorithm, which models the brain's own synaptic plasticity and incorporates local learning rules at each synapse.

While DNNs include fully connected layers and process continuous-valued inputs, SNNs are designed to be more biologically-realistic. Specifically, SNNs model neurons that have connections only within a local area of the entire network (or brain), and operate on discrete, time-based spiking events along synapses between said neurons.

The spikes are represented as binary electric impulses, and carry throughout the network all temporal and rate information associated with the neural model. Essentially, the more a neuron is activated, the more relevant it is in the network. There are no conventional weights, but, instead, spike trains; the spike trains maintain a trace of the moments and of the durations when neurons fire. The level of neural membrane potential at which a neuron spikes is computed with a specific differential equation. In the typical spiking model, a neuron's membrane potential is reset after firing, as this intuitively mimics brain functionality. This neural activity model, as well as alternatives, is described in the *Integrate-and-Fire Model* subsection, below.

To train an SNN, neuromorphic supervised learning methods perform worse than their supervised equivalents in conventional DNNs. Further neuroscientific discoveries will be required to establish a parallel between human brain functionality and how it can map to supervised SNN algorithms. Algorithms for unsupervised learning, however, such as Hebbian learning and the previously mentioned STDP, adequate for the current neuromorphic platforms, have shown performance nearing par with certain DNNs [2].

SNNs operate on spikes; therefore, inputted data needs to be obtained from a spiking source or the information must be encoded as spikes before entering the SNN. Correspondingly, the output from the SNN may be spikes or require conversion to non-spiking forms of information relay. The topic of converting data for the purposes of SNN input is explored in the following section.

### B. Generating Spikes

The preferable input to an SNN is an externally-generated spike train from a sensor or device that generates spikes. However, in many instances, especially when building an SNN from a CNN or DNN, creating spikes from other sources of non-spiking data is required [2]. A widely used mathematical method for this algorithmic generation of spikes is based on the concept of a Poisson process [2][3]. This describes a series of sequential events that occur at consistent average intervals but with only probabilistic, nondeterministic precise timings.

A vector or a matrix of continuous values - such as pixel values, frequencies, statistical data and financial trends - is inputted into an SNN by passing it through a Poisson process that defines specific sensibility thresholds and firing rates aligned to a Poisson distribution. A Poisson firing rate is the average number of spike arrivals per unit of time; it can be a constant, homogeneous spike rate, $\lambda$, or a time-dependent, non-homogeneous spike rate, $\lambda$(t). Strategies using Poisson processes to convert conventional data into spikes include rate-coding, temporal-coding and sparse temporal-coding [4][5].

Rate-coding is the process of converting DNN activation rates into spiking impulse trains [2]. In a DNN-to-SNN conversion the conventional weights are also replaced by attributes of the spikes, such as leak rates or refractory times. Networks like Boltzmann machines, described in a later subsection, and recurrent neural networks (RNNs) have also been translated to SNN versions using the leading approaches of the Neural Engineering Framework [30] and Diehl et al's method [6].

The main advantages of the rate-coding strategy used by these frameworks are ease of implementation when converting from DNNs, and nearing equivalent performance in classification tasks relative to associated DNNs.

With temporal coding, the earliest neuron spikes are considered when training the network. It is more computationally and energy efficient, due to the reduced number of spikes, and implements a function for avoiding false spiking through prior input decay [3].

To interpret the results from an SNN using conventional measures, reversing the conversion process used for the inputted data is required.

### C. Integrate-and-Fire Model

The integrate-and-fire (IF) neuron model, proposed in 1907 by Louis Lapicque, is one of the most used in SNN engineering due to its ease of mathematical expression and functional implementation. There are many variations, including exponential IF and the fractional-order leaky IF, among which the latter is more widely used.

The main difference between the non-leaky and leaky IF variants regards the mechanism of discharging a neuron's accumulator as data is inputted into the network. For the leaky IF model, the charge of the accumulator reduces rapidly, similarly to the decay in a real synaptic system, which operates on ion diffusion. Whereas for non-leaky IF, the accumulator is reset to zero only if it reaches the threshold, as explained in [7].

The general equation that describes the IF process is Equation 1:

$$I(t) = C_m \cdot \frac{dV_m(t)}{dt}$$

It represents the time derivative of the Law of Capacitance: $Q = C_m \cdot V_m$, where $V_m$ is the level of voltage at the membrane, and $C_m$, is the capacitance. As data is inputted, $V_m$ accumulates spikes. When the spike accumulation reaches a defined constant limit, $V_{threshold}$, a neuron fires. If the limit is not hit, then the voltage charge might never be released into an output spike, except if a timeframe or refractory period is introduced to the model.

The general equation that describes the leaky IF process is Equation 2:

$$I(t) = I_c + I_r$$

The current $I$ is, according to the Kirchhoff's circuit laws, the sum of two values: the capacitance $C$ and resistance $R$. By introducing a resistor, the model has more control over the current and the neuron's potential. $I_r = \frac{V}{R}$ and $I_c$ coincide with the previously mentioned I from the basic IF model. Therefore, Equation 2 can be rewritten as Equation 3:

$$I(t) = C_m \cdot \frac{dV_m(t)}{dt} + \frac{V_m(t)}{R_m}$$

Equation 3 can then be further reformulated to define the change of voltage at a specific moment $t$ related to the membrane's time constant, as the difference between the input current's intensity related to the membrane's resistance and the voltage applied at the same moment $t$. The result is Equation 4:

$$(C_m \cdot R_m) \cdot \frac{dV_m(t)}{dt} = R_m \cdot I(t) - V_m(t)$$

If the input current hits the limit, $I_{threshold} = \frac{V_{threshold}}{R_m}$, the neuron spikes, else it leaks, or decays the potential charge.

The input of spikes leading to a neuron firing through an IF model is diagrammed in Fig. 1.

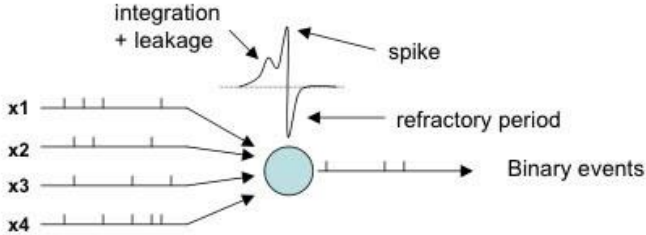

Fig. 1. The neuronal dynamics of an integrate-and-fire model can be represented as a summation or integration of input spikes. Source: [8].

### D. Converting Boltzmann Machines to SNNs

Boltzmann Machines (BMs) were among the first networks to be adapted to model an SNN. Proposed in 1985 by Geoffrey Hinton, a BM is an RNN with binary activation functions, which learns by minimising the error between the input data and the output with respect to a set of weights. Two modalities to use the Boltzmann machines for simulating a spiking neural network are described in [9] and [10].

A Restricted Boltzmann Machine (RBM) is characterized by having only two layers of neurons – one visible input and one hidden layer. Moreover, weights between the input and hidden layers are symmetric, there are no self-connected neurons, and neurons on the same layer are not interconnected. An RBM does not have memory. It analyses the inputted temporal, binary data and learns its probability distribution to generate predictions.

For modelling a spiking neural process on an RBM, [10] applied the Neural Engineering Framework (NEF), conceptualised on the basis of three main principles, as summarised by [11]. Firstly, neuron encoding must be nonlinear, while decoding must be linear. Secondly, any conversions or transformations of neurons must be applied to all neurons in a layer. Thirdly, control theory axioms can be applied to neuron state transitions. These principles are explained in practice in the *DNN Conversion for SNN Implementations* section.

A similar conversion of a BM to an SNN was performed by [9] but, in this instance, using Gibbs sampling, followed by a Markov chain-Monte Carlo method (McMC) that was coupled with a contrastive divergence algorithm. The training stage of this network consisted of two phases: In the first phase, Gibbs sampling generated values from the input and relayed them to the output layer. For the second phase, which acted like a form of backpropagation, the output ensemble of neurons reconstructed the input data with the McMC. The training stopped when the network reached equilibrium. To accelerate computation, an RBM was adopted instead of the classical BM and the training process adjusted accordingly.

Importantly, general Boltzmann models have limited power of computation for big data sets, as such the focus has been on DNNs.

### IV. NEUROMORPHIC HARDWARE AND SENSING

As encountered in the early days of neural networks, different variations of neuromorphic models are being rapidly created by researchers. Hence, from an implementation perspective, the boundaries between pure research tools for neuromorphic computing and tools for industrial adoption are still intertwined.

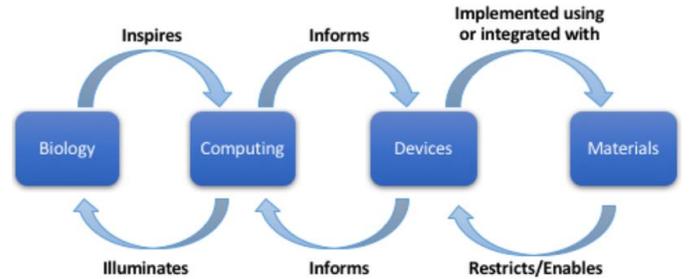

Fig. 2. Components of neuromorphic systems. The developments and tools of the parts are shared hence they influence the adoption of technology across the whole platform. This is common to most technologies that are more on the development and early adoption stage. Source: [12].

Fig. 2 illustrates the iterative development flow of neuromorphic systems, combining contributions from both hardware and software to implement SNN architectures.

### A. Hardware Architectures

Neuromorphic hardware can be broadly classified into digital, analog and mixed-signal circuitry. Several architectures have been proposed for implementing neurons and synapses in hardware.

Neuromorphic architectures are most commonly implemented using digital circuitry. The main advantages of this type of circuitry are the ease-of-development, low-power consumption and reusability. The major factor contributing to the preference of digital neuromorphic architectures over their analog counterparts is the low-cost associated with their development. One notable example is the Digital Neural Array (DNA), which utilises large-scale arrays of digital neurons. DNAs target both FPGAs and ASICs, depending on the application. For instance, FPGAs allow reprogrammability, while ASICs offer higher density and better performance despite the low flexibility.

Although more suitable for representing neuromorphic systems, analog circuitry is less represented than digital architectures. Analog implementations share many physical characteristics with neuromorphic architectures and, like SNNs, are robust to noise, making them an ideal option for hardware implementation. Such physical characteristics include reliability and asynchronous operation. The most suitable device for general analog circuits is the Field-Programmable Analog Array, or FPAA. A customisation of this device, specifically targeted towards neuromorphic applications, is the Field-Programmable Neural Array, which uses programmable components to mimic neuron and synapse functions.

A combination of the advantages of both digital and analog circuitry is achieved by mixed-signal implementations. In particular, the digital part of the system is exploited for fast processing and easy reconfigurability, and the analog part offers solutions to narrow bandwidth requirements. This makes mixed-signal devices particularly beneficial in communications systems [13].

Finally, there exist specialised SNN chips developed by industry firms and universities. Notable examples include TrueNorth developed by IBM, SpiNNaker from the University of Manchester, and Intel Corporation's Loihi processor. Each chip features unique characteristics, and the suitability of each depends on the desired application. It should be noted that all these SNN chips have been digitally implemented.

At a lower level of abstraction, these SNN-specific architectures are presently realised with traditional semiconductor devices, such as standard transistors on current technology nodes. However, future neuromorphic innovations will require new hardware paradigms to achieve brain-like performance. Memristors are an example of just such an advancement, coupling memory to individual transistors, mimicking neuronal memory in the brain [13].

### B. Neuromorphic Sensors: DVS

As a means of capturing data in a way that the mammalian brain does, one option is the utilisation of a Dynamic Vision Sensor (DVS), which is event-based and can provide an alternative mechanism to generate input signals for neuromorphic systems operating in real-time. DVS execution is based on relative changes in the temporal contrast of a given pixel, and the response of the sensor to these changes creates a series of spikes that can be directly read - i.e. without conversion and directly processed by a Spiking Neural Network.

A robotic application with neuromorphic hardware is presented in [14], which received input data read from a DVS sensor. The task for the robotic agent was to avoid obstacles by modelling an insect vision system. The retinal space - i.e. the lower half of the DVS aperture - is divided into three regions: left, centre and right. When one or more events are detected within one of these regions, a stimulus is sent to the robot, which responds by turning to the opposite side.

A method for training an event-driven classifier and converting it to an SNN is proposed in [15], as well as further explored in the *Present Neuromorphic Applications for MLDL* section. The approach compared synthetic and DVS data using both supervised and unsupervised-based learning methods. The authors argue that the use of synthetic input data - i.e. using Poisson encoding - demonstrates good classification results, whereas there is a drop in classification accuracy from data generated by a DVS neuromorphic vision sensor. However, a higher accuracy, 98.47%, was achieved using Fast-Poker-DVS data, a type of dataset characterised by a higher DVS resolution. Fig. 3 illustrates the results of classification accuracy using datasets with different sample sizes.

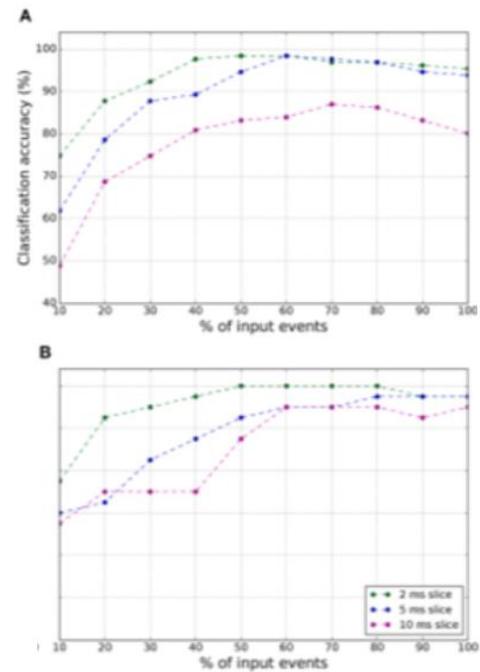

Fig. 3. SNN classification accuracy based on 131 (A) and 40 (B) samples from the Fast-Poker-DVS dataset. Source: [15].

### C. The SpiNNaker Computing Engine

One of the most notable existing computing engines designed specifically for neuromorphic systems, is the SpiNNaker (Spiking Neural Network Architecture), developed by the University of Manchester, UK. The main purpose of this massively-parallel, real-time distributed engine is to simulate the behaviour of large-scale Spiking Neural Networks (SNNs). Driving this massive-parallelism is the utilisation of multiple cores that form a single system, distributing the data over the cores to be processed in parallel with the lowest possible power consumption. This parallelism is a major characteristic of the biological functionality of the brain and is core to neuromorphic computing systems.

### D. Architecture of the SpiNNaker

SpiNNaker consists of up to 1,036,800 ARM9 cores, which communicate with each other through information packets that are transferred through a custom interconnect fabric. Multiple small packets can be transmitted simultaneously, with large bandwidths of more than 5 billion packets per second being achievable. About 40 or 72 bits are needed to represent one packet; these packets, transferred through the interconnect fabric, are based on a parallel hardware routing system. The responsibility of the router is to route the incoming packets for processing to other cores within a single node, or even to other nodes of the system. The cores are distributed over the system in around 57,600 nodes, and each of these nodes utilises 18 ARM9 cores, hence 1,036,800 cores in total. Furthermore, as depicted in Fig. 4, each node has Synchronous Dynamic Random-Access Memory (SDRAM) and peripheral support for Ethernet and GPIO. Out of the 18 cores within a node, typically 16 of them are used for information processing; the 17th is used for monitoring and supporting an operating system, and the 18th is backup in the event of a core failure. For a node to be considered viable, at least 17 of the cores must be functional. A characteristic of SpiNNaker is its multiple levels of abstraction for each component, even given its extremely-large size, enabling low-effort fixes to faulty components. The connection between the nodes is achieved with the use of six inter-chip links. Assuming that 4-bit symbols are transmitted at 60 Hz per link, then $6*10^6$ packets per second can be processed; hence, with six such links, the processing speed increases to $3.6*10^7$ packets/second [16].

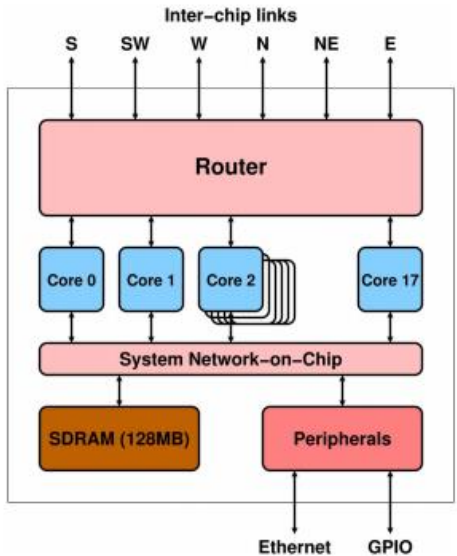

Fig. 4. Components of a SpiNNaker node. Source: [17].

### E. SpiNNaker Software

Software can either run on SpiNNaker directly and/or on other machines that interact with SpiNNaker. These can be classified into control software and application software. The former contains bootstrap code that provides fundamental services, such as loading code through either the inter-chip links or the Ethernet. This software is designed to run on one of the 18 cores within a node to allow the application to execute on the remaining cores; this is known as the SpiNNaker Control and Monitor Program (SC&MP). Application software essentially includes the instructions to be executed in SpiNNaker, and it is usually written in the C language. The host workstation is the system that controls the operation of the entire SpiNNaker system and typically runs on Linux OS. Fig. 5 is a diagram of the host system's interplay with the entire SpiNNaker chip.

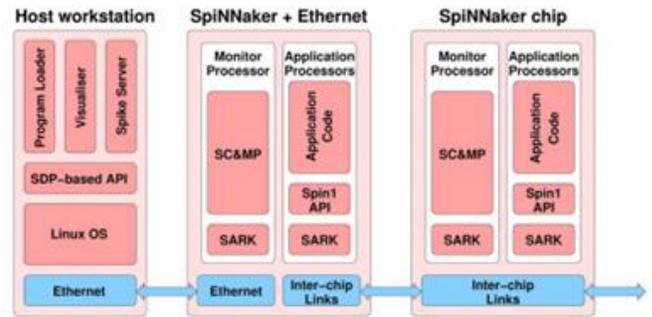

Fig. 5. Software component interface between host machine, root node and rest of the SpiNNaker nodes. Source: [17].

Applications essential for neuromorphic computing include the spike server, which produces spikes in real-time during a simulation in SpiNNaker, and the visualiser, which allows visualising the spikes at the output of a particular section of the system. In addition, Ethernet networking technology is used for communication with the root node and the various other nodes of SpiNNaker [17].

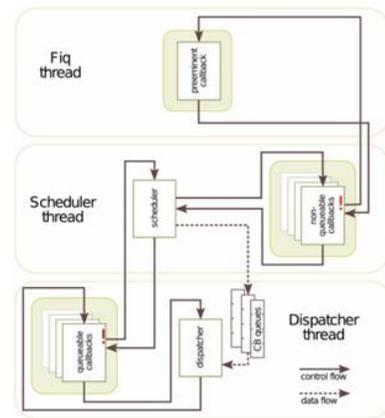

Fig. 6. Framework of the interrupt-driven SpiNNaker software. Source: [17].

The software model of SpiNNaker, shown in Fig. 6, is event-driven with real-time execution. This means that the system or a part of the system, such as a single node, will process information only when an interrupt event occurs. Outside of these interrupts, the system remains in a sleep mode, thus consuming low power for the majority of its operation time, assuming that events are infrequent relative to total runtime. If more than one event occurs at the same time, then the scheduler has the responsibility of scheduling the interrupts by placing them in queueable callbacks. A non-queueable callback can be defined as the foremost callback in a system, having the highest priority and being able to preempt other callbacks, which can be either queueable or non-queueable. This is achieved using an Application Programming Interface (API) for abstracting the complexity of the interrupt system from application developers [17].

## V. STATE OF NEUROMORPHIC COMPUTING

Technologies based on SNNs are still in the early stages of industrial adoption, however, the core algorithms are still evolving. From a user perspective, this divides the neuromorphic digital ecosystem into the broad categories of model development tools for neuroscience research and algorithms or tools for industrial applications. In the previous section, the underlying neuromorphic hardware was examined. The following section delves into the digital tools available for this hardware and the current state of neuromorphic technology adoption.

For neuroscience research, the digital simulators and tools that are available are generally used to simulate both a microscopic and macroscopic view of the brain. They are used by neuroscientists to easily digitally model the brain and test new theories regarding the functioning of the brain. This involves the neuron-level model and the network-level model [18]. The neuron-level modeling involves simulation of the neuronal physiological behaviour in terms of electronic components, such as resistors, capacitors and inductors. Practically, this aided in the development of the pure analog devices that were mentioned in the previous section. The network-level model involves algorithms and tools that control the learning rate, as well as rules that govern the connection between thousands of neurons and how input events are translated. Some of the popular simulators, such as CARLSIM [19] or HRLSIM, which use existing GPU or CPU processors, provide realistic analysis of larger networks. Defining and handling connections between thousands of neurons in biological time is a fairly complex task, though.

Other simulators include NEST, BRIAN and ANARCHY [20], which are geared towards the neuro-biophysics community. PAX [18] is another popular tool that's both a hardware and software simulator. Depending on the architecture, these simulators also have digital hardware tools that connect the network-level algorithms to the hardware implementation of the neuron model. BRIAN is an open-source option and has a Python backend with Internet browser capabilities for rapid experimentation. For the ML community, these systems might require familiarisation with biological syntax. Hence, BindsNET [20], which was built atop the PyTorch deep learning library that includes TensorFlow and SpiNNaker support, should be of interest to ML practitioners. Nengo is a similar framework that is explained in detail in the upcoming sections.

As depicted in Fig. 7, greater mimicking of the biological properties of the neuron results in an increase in computational load and further complicates translation to a hardware or software model. To transform these research ideas to be industry-ready, a compromise must be reached. Industry giants, such as Intel and IBM, have provided their own realisation based on where their expertise falls on the spectrum between hardware accuracy and software implementation. This was made possible by either building onto their proprietary hardware architectures or, like SpiNNaker, working with existing hardware architectures and developing their own software platforms.

Therefore, based on different applications, a combination of neuromorphic software and hardware can be used with varying levels of abstraction. A typical design flow of a neuromorphic system is diagrammed in Fig. 8, below. Stage 1 comprises platforms like MATLAB, Octave, TensorFlow, as well as simulators, such as CARLSIM, HRLSIM or BindsNET, which are typically used to test new conceptual models. They are used to define the network-level models, and thus the learning rules mentioned earlier.

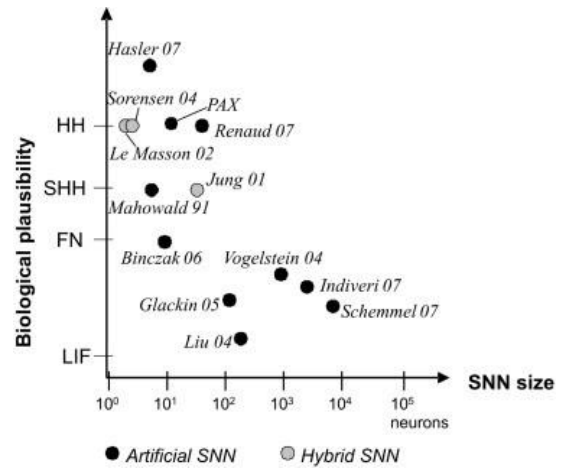

Fig. 7. The more complex models that closely mimic the biology of the brain have more computation load on the underlying hardware. Compromise between computational costs, hardware complexity and network size are required for a real time implementation. Source [18].

Stage 2 are hybrid systems, like Nengo, BRIAN and PyNN, which are typically used in simulation and small-scale testing applications. They can perform conversion from existing NNs to SNNs for faster implementations.

Finally, stage 3 systems use hardware platforms like Loihi, TrueNorth [21], and SpiNNaker. Also included in this stage are hardware synthesis tools, which can create custom, software-defined hardware models and enable control over neuron-level models.

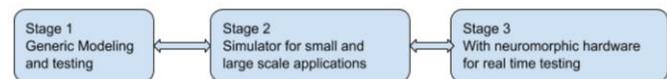

Fig. 8. The stages in a typical neuromorphic model design flow. The arrows indicate the interdependency and flexibility of the design stages. Not all stages are required and are highly dependent on the intended application.

As the arrows in Fig. 8 indicate, the distinction between stages is quite fluid, and tools of each stage can be used interchangeably depending on intended applications - as examples, pure research, testing or future deployment. Presently, "NeuCube" [22] is one of the first generic system architectures that supports all 3 stages. It is based on SNN advances, and supports languages like MATLAB, Java, Python with PyNN, and C++. NeuCube is also available for computational platforms such as SpiNNaker, GPUs, and cloud-based deployments. It is primarily being developed to model the brain, and will be useful for future innovators who wish to develop tools that will work at the intersection of reading the brain and creating AI systems that can interact with the brain directly.

Some of these software tools provide complete abstraction from the layers of model complexity we mentioned above. This opens up development to non-experts: the developer community who are ultimately responsible for the adoption of these technologies in real-world applications. The simpler these tools are to adopt; the wider will be their acceptance. The easiest way to achieve this is to merge SNN with existing NN platforms that have been built and tested over decades of research.

In the coming section, such an easy-adoption approach is investigated, leveraging mature and stable MLDL toolkits - TensorFlow and Keras. Concepts facilitating the effort of MLDL practitioners in connecting existing NN technologies to future SNN opportunities are also explained.

## VI. DNN CONVERSION FOR SNN IMPLEMENTATION

While SNNs may be better suited for real-time, event-based stimuli, they can also bring benefit to recorded image and video challenges currently undertaken by convolutional NNs (CNNs). The conversion of conventional artificial NN (ANN) and DNN models into SNN implementations provides an incremental path for machine learning engineers and specialists to transition from continuous-valued datasets to spiking, event-based ones. As mentioned in the *Fundamentals of SNNs - Generating Spikes* subsection, above, the transformation of traditional neurons into spiking variants can be performed through either encoding the activation functions of artificial neurons into analogous spike transfer functions in SNNs, or translating temporal information into SNN-compatible spike trains.

### A. Rate Conversion

Rectified Linear Unit (ReLU) layers are one of the standard activation functions in ANN and DNN models. Along with sigmoid and hyperbolic tangent activations, the ReLU function provides a stepwise gradient with which to differentiate results. The process curve of a spike triggering through receiving sufficient current to fire can be approximated as the zero-or-one step-function characteristic of a ReLU; thus, a mapping between SNNs and DNNs can be established. Other conventional layers, such as Max-Pooling, Softmax and Batched Normalisation do not have SNN equivalents, however [3].

The usual assumption is to directly convert one DNN neuron to one SNN neuron; however, it is possible and efficient, from a resource utilisation perspective, to map multiple SNN neurons to one analog neuron. Moreover, unlike ANN and DNN models that must complete a minimum of a single epoch of learning to derive any meaningful results, the SNN implementation of a conventional model can derive results almost immediately from the first layer of spiking neurons [3]. Of course, similar to DNNs, deeper SNN models that are left to run longer will produce more accurate results. Nevertheless, this ability to obtain results quicker than traditional DNNs, albeit at lower accuracies, is an enticing benefit of SNNs for specific, real-time applications.

Given the complexity of DNN models, approximations of artificial neurons to spiking versions are bound to result in errors and potentially inefficient translations. Applying weight regularisation has been proposed to prevent an exploding gradient-style problem with errors building up in networks converted to SNNs [6]. In addition, applying dynamic firing thresholds for spiking neurons [23] and introducing artificial noise into SNNs [24] have been suggested as mechanisms to both temper the number of SNN neurons required to convert an ANN or DNN, as well as improve the overall performance of the resulting model [3]. Ultimately, simpler, binary neural networks (BNNs) have fared better during rate-encoded transformations to SNNs. The weight quantisation of BNN neurons already closely approximates the on-off firing of spikes, with no spikes required in the case of translations from zero-weighted BNN neurons.

### B. Temporal Conversion

Spike-timing-dependent plasticity (STDP) with non-leaky integrate-and-fire (IF) neurons most closely resembles the mechanism used by an actual brain to generate spikes [25]. This method is non-differentiable, however, and the principles of backpropagation of errors used in conventional stochastic gradient descent (SGD) cannot be applied. The rate-based conversion approach previously described sought to overcome this hurdle by swamping activation functions. A time-based scheme whereby errors can be distributed back through SNN layers is proposed in [26].

Unlike traditional ANNs, which have no memory attached to their neurons, SNN neurons, like those in an actual brain, can retain state and carry forward actions of spikes through layers. Spike LAYer Error Reassignment (SLAYER) has received significant attention as a solution to differentiating the spiking function. Through SLAYER, axonal delays and synaptic weights are learnt during training, with the "credit" of errors being back-assigned through layers [26]. These credits are calculated as the number of true and false spikes inputted to a neuron. If the number of positive spikes is greater than negative ones, then the neuron triggers a spike. Not only does this provide a derivation to the spiking function as the probability density function of a neuron spiking, but it also associates a representation to no spiking events as learnable false spikes – solving the "dead neuron" problem [26].

### C. Asynchronous Model

Another, spiking-like technique, which replaces the continuous clock in traditional systems with a pulse-based, or "click" approach to circuit timing has been hypothesised as an alternative to spikes. These asynchronous CNNs (ACNNs) theoretically benefit from the same energy and resource efficiencies provided from SNNs, as their elements are only activated when required – similar to spikes –, but also promise to directly leverage existing DNN models with minimal to no conversion required [27]. While presenting an even simpler path for conventional ANN and DNN models to be implemented using energy-restrained devices, as of time of writing, further research must be conducted to properly contrast ACNNs and SNNs. Without an ACNN model to test for the research performed for this paper, a DNN to SNN conversion was solely tested. Additional experimentation on ACNNs is to be conducted as part of future investigations.

One of the deepest, most complicated DNNs to be successfully converted to an SNN is AlexNet [24]. This was performed using the Nengo toolkit, which was selected for the real-world experimental setup described later in this paper. Also, [3] observed exemplary results in converting VGG-16, a popular CNN, into SNN form, with integrate-and-fire neurons, which were previously described. The success of [3] with VGG-16 influenced the use of this model for the follow-on experiment using Nengo.

## VII. RAPID SNN DEVELOPMENT WITH NENGO

Nengo is a pioneering SNN toolkit released and supported by Applied Brain Research. It is based on the neuroscience-adhering Neural Engineering Framework (NEF), developed by Eliasmith and Hunsberger [24], and is credited as the framework powering the cognitive potential of the world's most complex brain model, Spaun [30].

The neuron activity of the human brain is modeled in Nengo as a network of connections between "ensembles" and "nodes". A group of neurons, which, for Nengo, are real-valued state vectors that change with time, comprise an ensemble, whereas nodes represent stimuli to the network. The connections between nodes and ensembles are captured as weight matrices, which encode the strength of the bonds between neurons [30].

With Nengo, an SNN can be defined and implemented from low-level constructs using networks of ensembles and nodes, or an existing DNN can be efficiently converted to a rate-based SNN using a converter method built into Nengo models.

Nengo was selected to rapidly implement an SNN model for this paper. Given its integration with the Keras API and TensorFlow backend, the learning curve for an ML practitioner to ramp up on Nengo is less steep than implementing SNNs from fundamental principles. The platform also supports direct integration with a number of leading neuromorphic hardware processors, including the previously detailed SpiNNaker, the Intel Loihi chip, and FPGA SNN accelerators.

### A. Nengo CNN Translation

Nengo applies rate-based transformations to quickly convert an existing DNN to an SNN form that can be simulated. As described earlier in the paper, rate-based encoding translates maximum magnitudes from ReLU layers in an DNN to groupings of spikes from IF neurons in an SNN [28]. This is the most widely used conversion scheme, first proposed by [6].

The process of batch normalisation is frequently leveraged by DNNs to improve generalization. This methodology introduces neural biases, which cannot be translated to SNN neurons, and, hence, batch normalisation layers are removed by Nengo [28]. To compensate for the loss of regularisation afforded by batch normalisation, neuronal weight dropout layers are inserted in their stead. It was found by [29] that dropout layers should be substituted for all batch normalisation, except in instances where a pooling operation would follow the batch normalisation.

As spikes are quantised as either on or off, or 0 or 1, max-pooling operations, which are effective at reducing feature complexity between layers in conventional ANNs and DNNs, need to be swapped for average pooling layers prior to SNN conversion. These layer substitutions are managed through Nengo converter settings [30].

Since ReLU weights and activations from an ANN are converted to a time-series of spikes, it is critical that the DNN weights are judiciously generated. Further findings from [29] suggest that placing average pooling layers ahead of ReLU functions in a DNN-to-be-converted results in an improved performance accuracy while minimally impacting latency.

### B. Model and Dataset

The Modified National Institute of Standards and Technology (MNIST) dataset of handwritten digits, created by LeCun et al, is a standard benchmark in evaluating computer vision models [31]. Building from this well-reputed foundation, Orchard et al developed a Neuromorphic MNIST, or N-MNIST, by converting the static MNIST digits into spiking impulses. Their work followed on from the MNIST-DVS dataset, which used a DVS camera pointed at a screen with moving MNIST objects to generate spiking representations of these otherwise static images [32].

For their N-MNIST dataset, Orchard et al were inspired from human rapid eye movements, known as saccades, which are instrumental in the biological reception of both static and dynamic objects into the brain.

To mimic saccadic movement, Orchard et al generated a series of spiking events for each MNIST digit by shifting and tilting the DVS sensor, focused on unmoving images. This differs from MNIST-DVS, a prior dataset, where the sensor was fixed but the images moved. The results from [32] suggest that moving the DVS instead of the images results in more biologically plausible spike generation.

The resulting series of spiking events is then a representation of the pixel brightness intensities morphing through time as points on the MNIST images are captured by a DVS from different angles [32]. The N-MNIST dataset contains the sequence of spiking events for each traditional MNIST digit, along with the x- and y-axis coordinates at which the spiking events were captured. Just as MNIST is now a de facto comparison benchmark for object detection and classification algorithms, so too can N-MNIST become one of the benchmarks for SNN implementations. For the experimental results that follow, however, the simple MNIST dataset was used to showcase the relative ease in porting both conventional datasets and models to SNN form.

VGG-16 is one of the foremost computer vision CNN models. It has a suitable complexity, featuring 16 convolutional and fully connected layers, for rigorous object classification tasks, while remaining sufficiently compact for conversion to an SNN using Nengo on restricted computational resources.

The composition of VGG-16 as predominantly convolutional and dense DNN layers lends well to the layer substitutions described in the prior subsection. Moreover, one additional alteration was performed based on results from [29]: One of the fully connected layers was removed, as this proved to reduce noise and perform better in this study. The modified "VGG-15" along with a stock MNIST dataset was translated to an SNN model using Nengo.

## C. Nengo Toolchain

Nengo runs Python at its core, hence most of the existing Python tools for visualisation and computation can be directly accessed through Nengo. For the experimental implementation, a mix of GUI and direct programming approaches are possible. Nengo Core offers the flexibility to build various neuromorphic architectures for different applications. Additionally, ML practitioners already familiar with TensorFlow can quickly learn to use the NengoDL wrapper to write TensorFlow-like models and easily port existing code to Nengo. To gain a deeper understanding of Nengo and NengoDL, a barebones SNN was implemented from a conversion from the aforementioned, modified VGG-15 model without any hyperparameter optimisation applied.

## D. Experiment Setup

Careful consideration for available system memory was required in order to successfully complete the CNN conversion to SNN using the NengoDL wrapper. A small mini-batch size of 64 was found to be necessary so as to not overrun the available 10 GB of system RAM.

many instances where CNNs are already used in the field of MLDL and beyond. Nengo can directly leverage existing code and models from popular ML toolkits, such as TensorFlow and PyTorch, permitting a direct path for algorithm migration and quick prototyping of results.

The power-saving and real-time execution of SNNs position these models as evolutionary next steps for some conventional ANN and DNN applications. In addition, there are numerous use-cases for SNNs for which no analogue in conventional NNs exists. These constitute the most interesting applications for SNNs, as they make possible new technologies and discoveries previously only imagined.

Key current applications for SNNs are described in the following subsections, providing motivation for ML engineers and data scientists to explore SNN implementations for their algorithms and projects. These applications showcase the power reduction advantage, on-fly learning and quick response to external stimuli of neuromorphic processing. The future potential for SNNs and neuromorphic sensing and processing technologies is then overviewed in the following section, providing a glimpse into exciting new realms for MLDL and AI as a whole.

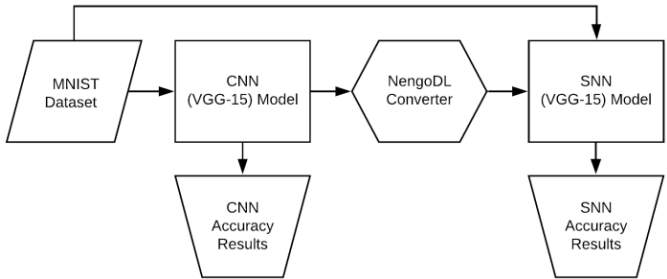

Fig. 9. Block diagram representation of the experiment setup showcasing rapid conversion of a VGG-15 CNN to SNN form using NengoDL.

A diagram overviewing the experiment setup can be referenced in Fig. 9. The VGG-15 model, described in the prior subsection, was first trained with the MNIST dataset. The non-spiking accuracy of this model was found to be 98.75%.

An SNN conversion was then performed using NengoDL, and the new model was then revalidated against the same MNIST dataset. Using a spike firing rate of 300 Hz, an accuracy of 88.75% was achieved. The performance of the model was determined to be highly contingent on the spiking rate. Setting the rate of spiking to 250 Hz, for example, resulted in a diminished accuracy of 68%. A plot of spiking results for four MNIST digits can be reviewed in Fig. 10.

The goal of this experiment was primarily to showcase a real-world example of converting a popular computer vision algorithm to SNN. Further spiking rate and regularisation optimisations will be required for the accuracy of the converted SNN model to be on par with the original VGG-15.

## VIII. PRESENT NEUROMORPHIC APPLICATIONS IN MLDL

Neuromorphic computing can be utilised in many machine learning and deep learning applications, some of which are described in this section. Converting an existing CNN to an SNN using Nengo showcased the ease with which ML specialists can begin immediately observing the advantages of lower power and reduced latency that can be achieved by running their algorithms through spiking. SNNs can be used in

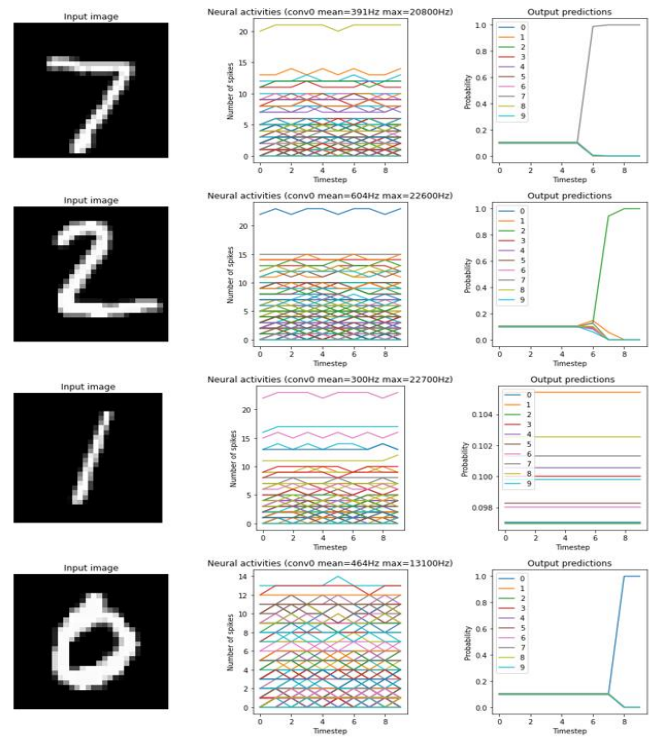

Fig. 10. MNIST digit predictions using the Nengo-converted VGG-15 model to SNN. The central column demonstrates spiking activity over ten timesteps. The output prediction probabilities from the SNN model, right-most column, clearly show the confidence levels for each of the MNIST digits. Images for "7", "2" and "0" were well-predicted, with ambiguity remaining in the identification of the handwritten "1".

## A. Pattern Recognition

The use of SNNs enables quick and adaptive pattern recognition. A suitable SNN architecture is proposed by [33], which consists of integrate-and-fire neurons in a hierarchical arrangement within a network of four layers. The network is designed in a way that the last neuron collects information from previous neurons to make a final decision. In this work, facial recognition is the specific form of pattern recognition examined. Multiple points of view are considered for recognising faces. As the authors state, the main advantages of the neurons in an SNN are the low computational complexity and the hierarchy of the spikes at each layer; i.e., the first presynaptic spikes have a higher weight and hence contribute more to the final outcome as compared to the remaining spikes. Fig. 11 illustrates the SNN architecture utilised for pattern recognition, with each of the four layers having a specific function. The first layer enhances the high-contrast regions of an input image; the second layer calculates the orientation of selective cells; the third one is the first training layer, in which neuron maps are formed; and, the last layer forms a prediction based on information from the previous layers.

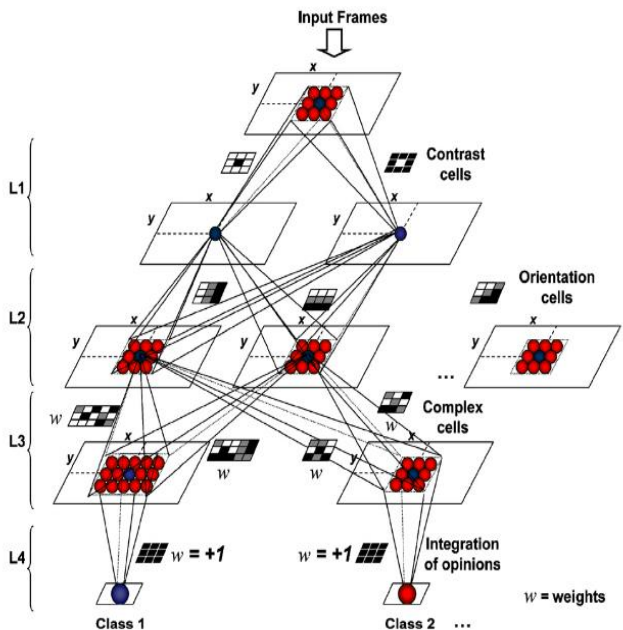

Fig. 11. SNN architecture with four layers. Source: [33].

### B. Classification

An SNN-based classifier, previously explored in the *Neuromorphic Sensors: DVS* section, is another common application in neuromorphic computing. In [15], an event-driven classifier is proposed. A non-fully connected Spiking Convolutional Neural Network (SCNN) is used to extract high-level features from input images. After the features have been extracted, the output of the SCNN is connected to the input of an SNN, which will be used for training. It should be noted that in [15], the topology of the SNN is fully-connected as opposed to the topology of the SCNN in the previous stage of the system. The methodology used is unsupervised and two hidden layers were specified for training. Fig. 12 illustrates the aforementioned topology.

### C. Early-Event Prediction

An application of early-event prediction is presented in [34]. It has been proved that SNNs are preferred over CNNs for learning spatio-temporal patterns from spatio- and spectro-temporal data (SSTD). SNNs primarily learn and predict patterns through temporal encoding of data. In [34], static and dynamic temporal variables describe an individual vector $x$, and the nearest samples to that vector, in the representation space, are obtained by measuring the distance between them. The outcome for $x$ is predicted from the data provided by the closest neighbouring individuals to $x$, and is described by the probability of an event happening and the accuracy of its occurrence. An iterative method is then performed for optimising the outcome prediction.

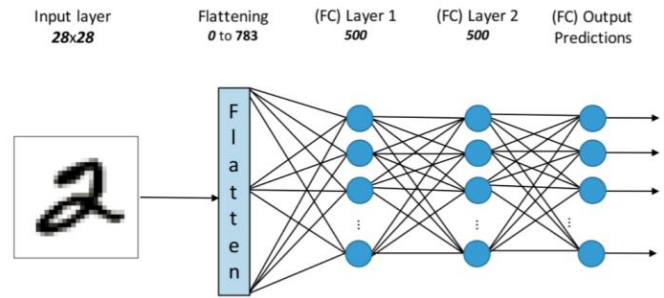

Fig. 12. SNN topology for the event-based classifier. Source: [15].

## IX. Future Neuromorphic Applications

### A. Human-Computer Interface

The Brain Computer Interface (BCI) developed in the past decade has resulted in numerous innovations across various fields, including psychology, healthcare and even everyday consumer products. BCI heralds an era where human thoughts can become action without muscle movements. Initially, BCIs were mostly developed by the research community for medical applications, especially for people who have lost portions of their motor abilities. Increasingly, they are now being developed for industrial purposes, everyday applications and casual users, as summarised in Fig 13.

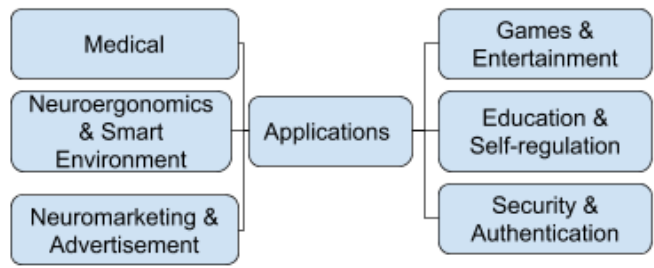

Fig. 13. Some traditional and futuristic applications of BCI. Source adapted from [35].

Until recently, most BCI innovations centered around applications that regarded the brain as a black-box and analysed outcomes from inputted stimuli. SNN developments, with their biologically-mimicking spiking nature, have the potential to revolutionise BCI. It will soon be possible to apply BCI to a brain both *in vitro* and *in situ*, while a person's brain is generating signals. This opens up the possibility of two-way BCI communications, as such a model can then interact with the brain in real-time.

Prior BCI research has focused on communications solely through neural feedback. Future innovations promise universal translator-like, brain-to-brain communication capabilities on the order of a "babel fish" [22], previously considered purely science fiction. The objective is to empower a brain to communicate with both other human brains and artificial general intelligence systems, described in the next section, in real-time.

Current BCI implementations are synchronous in nature and require system initiation, so interactions are possible only for fixed time windows. These interfaces count event-based potentials that are triggered by stimuli, such as sounds or visuals generated for predetermined timeframes. Interested readers can reference [35] for an example of such a device, the P300 Speller.

BI, or bio-inspired BCI leverage the asynchronous property of SNNs, which can be directly initiated by the user outside of fixed time periods. They are always available and represent a more biologically-plausible method of communication [35]. Whereas present synchronous systems also encounter problems distinguishing real and imagined actions, BI-BCI have proven to be better at detecting and recognising such scenarios. Nevertheless, further research and developments are required in the hardware domain to build devices that will be fully neuromorphic and more suitable to support BI-BCI technologies.

It is possible to envision a future where people will be able to communicate with AI devices and other humans, brain-to-brain, as well as interact with system that could archive replicas of an individual person's thought patterns. Some technologists, like Moran Cerf [36], are inventing devices that can even read dreams. Perhaps, it will one day be possible to upload a full representation of a human's brain to a data centre cloud, and bio-inspired SNNs are paving the path for this to become reality.

### B. Multi-Sensory Input

The fusion of different sensory systems for SNN applications is a necessity that began to be addressed with individual projects such as DVS and AEREAR2, for vision and auditory senses, respectively. DVS was previously explored in the *Neuromorphic Hardware and Sensing* section. AEREAR2 was created in 2016 by Shih-Chii Liu and her team, who presented a silicon cochlea prototype, 10.5 x 4.8 millimetre-square ($mm^2$), with a 0.18 micrometre ($\mu$m) 1P6M CMOS chip and its core system operating at under 0.5 V. The device has two, 64-channel modules - around 15 times less than the number of channels of a human ear - for the human-audible frequency range of 20 Hz and 20 kHz. It can an output 100,000 events per second. The silicon cochlea uses Automatic Power Factor Control (APFC) circuitry and models the cilia and spiral ganglion cells.

Combining two or more receptors increases considerably the intelligence of a technical system. For an audio-visual module, should one of the two sensors malfunction for a specific task, the system can dynamically adapt and rely on the other module in order to deliver a solution; for example, computing the distance from the stimulus to the receptor, either by analysing the sound duration or intensity, or by analysing the object's displacement in a scene over a series of images. This adaptive leveraging of sensors well-replicates the human brain's own adaptability in reacting to the availability or loss of various senses.

A second motivation for integrating different sensors, in particular for audio-visual applications, is in the combined potential of these systems working together to cross-check findings and produce results greater than the sum of their individual sense receptors. Again, this models the human capacity to feel, understand and react to the physical world using all five basic senses simultaneously.

A multi-modal, fusion neuromorphic sensor was implemented in 2012 [37] for source localisation purposes using a pair of 32-channel silicon ears and a 40x40-pixel silicon vision sensor. Such a multi-sensory system coupled with SNN interfacing has the potential to accelerate research both in biological human interactions, as well as in the advancement of BCI applications, as previously explored, and autonomous humanoid robotics.

For continued research, additional commercialised neuromorphic audio-visual sensors include DVS128, PAER and DAS1 [38].

### C. Artificial General Intelligence

The algorithms and learning models so far described and tested, for both DNNs and SNNs, have concerned specific applications or tasks. For example, the DNN-to-SNN conversion performed as part of the experiment in section *Rapid SNN Development with Nengo* pertained to classifying digits in the MNIST dataset. This same model could not then be applied to other tasks, however. This specialisation is known as "narrow AI" [39].

The human brain is a generalist system that can rapidly adapt to previously unexperienced or unforeseen situations and circumstances. The goal for future developments in AI and MLDL is to achieve equivalent and superior levels of generalisation potential for computational learning systems.

Whereas generalisation of a dataset is sought for ANN and DNN algorithms, these are focused on specific problem-sets and domains, at which they can achieve exceptional levels of accuracy. These algorithms are modelled on the brain, but do not function in the same event-based, spiking manner. SNNs, as has been explained, are based on a spiking, and can more readily interface with the brain [40]. Combining both DNN and SNNs approaches across the range of problem-sets, carrying forward knowledge from each to apply to the next, or simultaneously in parallel, is the objective of Artificial General Intelligence (AGI).

Similar to the development of novel hardware to most efficiently power SNNs and neuromorphic devices, processors that can both effectively compute DNN and SNN algorithms will be required for AGI to perform at levels comparable to narrow algorithms. Merging the full-precision weights and matrix multiplications required for ANNs and DNNs with the binary spiking time-series used for SNNs is the challenge facing AGI hardware designers. One existing prototype processor, Tianjic, is capable of unifying DNN and SNN operations on a single chip, and has been shown to be viable through a fully autonomous bicycle proof-of-concept [40].

Those researching AGI seek inspiration from human development in empowering learning systems to generalise and learn on their own through building on foundational knowledge. The most direct path to enabling AGI is through the amalgamation of existing specific, narrow learning models, in much the same way as the human brain is a composite of various sectors responsible for different tasks [39].

AGI platforms, such as SingularityNet and OpenAI Universe, aim to train learning models using numerous, real-world applications requiring a variety of skills that must be learned. SingularityNet performs this through a blockchain of learned computational experiences, creating a distributed mind approximation [41]. Universe by OpenAI, on the other hand, is leveraging the action-reward mechanisms underpinning reinforcement learning to train a collection of algorithms to excel at video game playing, online navigation and other common human-computer interfacing actions [42].

As the adaptive knowledge stored by advanced AGI systems increases, SNNs will be required to both simulate a spiking brain interface, as well as maintain the low-power, low-latency and parallel execution of these increasingly complex tasks.

## X. Conclusion

Current software development frameworks for MLDL, such as Nengo, and hardware platforms, like SpiNNaker, are evolutionary steps towards a wider adoption and proliferation of neuromorphic algorithms, processors, and sensors. These systems lower the SNN learning curve and empower data and machine learning scientists and engineers to immediately begin migrating existing solutions and obtain direct results from SNN implementations of their tools and applications. With an increasing preponderance of MLDL experts researching, working on and popularising SNN and neuromorphic advances, a shift towards these highly parallel, low-latency and -power algorithms, for the applications in which they excel, will occur.

As with ANNs and DNNs, the mass adoption of these machine learning paradigms was propelled by strong community involvement in open source toolkits, as well as the introduction of leading players in the technology domain, such as Amazon, Facebook, Google and Microsoft investing heavily in the latest research and spurring increased attention from programmers and developers from around the globe. Similar investments and an outpouring of development community support will be required to elevate SNN models and related neuromorphic devices into the MLDL spotlight.

Nengo is one of a number of toolkits that are facilitating developer onboarding onto SNNs. As has been demonstrated through the experiment detailed above, extending an existing, widely known and powerful CNN to leverage the benefits of SNN and operate on neuromorphic problem-sets is a minimal overhead process. Further developer community and industry involvement is required to lift the level of maturity of these SNN tools and development environments to that of modern CNN and DNN frameworks, such as PyTorch and Tensorflow. The building blocks are all present, though, for a revolutionary shift in MLDL to occur and for the community to develop and deploy SNNs for autonomous and remote-sensed applications in devices, in the cloud and at the edge.

This paper serves to educate MLDL professionals and amateurs alike on SNN theory, the neuromorphic hardware that is currently available, and the software platforms that abound to rapidly get started and ramp up. It is intended to stimulate researchers and industry decision-makers to investigate SNN and neuromorphic technology for their MLDL projects, as well as catalyse the development community to rally behind SNN platforms. Ultimately, it acts as a checkpoint for the present state of neuromorphic and the promising directions in which this rapidly accelerating field is heading.

## Acknowledgments

We wish to thank Prof. Deepak Uttamchandani for his project guidance and teamwork support throughout the course of this research. His constructive feedback and coaching ensured that we achieved our desired objectives.

We received valuable insight into the foundational theory and future potential of neuromorphic technologies and SNNs from Prof. John Soraghan and Dr. Gaetano Di Caterina, to both of whom we are grateful.

For explaining the biological analogies for SNNs, discussing the current research streams in the neuromorphic domain, and inspiring the Nengo CNN-to-SNN conversion experiment detailed in this paper, we would also want to express our appreciation to Paul Kirkland and Yannan Xing.

The depth of knowledge, experience and patience exhibited by those mentioned facilitated this research and enabled the results achieved.

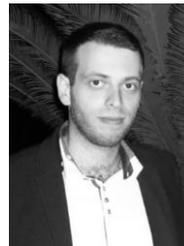

**Spyridon Bizmpikis** received his BEng (Hons) in Electronic and Electrical Engineering with specialisation in digital systems in 2019 from Glasgow Caledonian University, Glasgow, UK. In 2020, he enrolled for an MSc in Electronic and Electrical Engineering at the University of Strathclyde, Glasgow, UK, which he is currently completing. His interests lie in signal processing, including image and video processing and embedded systems design.

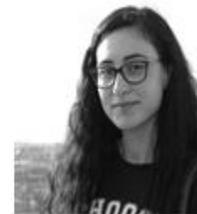

**Ionela-Ancuța Cîrjilă** is a computer science teacher in Constanta, Romania. She was awarded a BSc in informatics by the Faculty of Mathematics and Informatics from "Ovidius" University of Constanta in 2015 and a MSc in software engineering by the Faculty of Mathematics and Computer Science from University of Bucharest in 2017. Currently she continues her studies to deepen and broaden her knowledge in informatics with a master course in machine learning and deep learning at University of Strathclyde, United Kingdom.

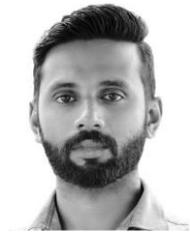

**Geet Rose Jose** received his BTech in electronics and communication engineering from Mahatma Gandhi University in Kerala. For the past 5 years he has been working with various startups in India to promote local innovations and is currently pursuing an Msc in Electronics and Electrical engineering with focus on Signal Processing and Machine Learning.

After his BTech he and his colleagues co-founded Resnova Technologies that focuses on solving local problems through indigenous innovations and research. His efforts in the past decade to build rural technologies has earned him numerous awards in his country including Young innovator award 2014 and 2015. He was honoured in 2015 with the Top Technical achievement award for these achievements during his undergrad. Over the next few years their startup gained recognition in various national and international forums and was recognised as one of the top 40 hardware startups in India in 2018.

His chief motivation is to build affordable technologies for the community. He pursues technologies that will bring down the cost of R&D and supports open source innovations.

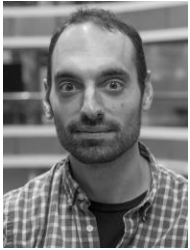

**Philippe Reiter** was awarded a BEng in Computer Engineering with specialisation in Biotechnology in 2006 from McGill University in Montreal, Canada. Following over a decade of industry experience, he crossed the pond and is currently completing an MSc in Machine Learning and Deep Learning from the University of Strathclyde in Glasgow, UK.

Consequent to receiving his BEng, he joined Advanced Micro Devices, Inc. (AMD) in Toronto, Canada as an IP and SoC design verification (DV) engineer in 2007. By 2014, he was a technical lead and had established and trained DV teams in eight of AMD's worldwide offices and offsite development centres. He next transitioned to business communications and spearheaded the modernisation and adoption of AMD's global intranet from 2014 to 2019. With a new digital workplace strategy fully implemented and deployed, and AMD teams across the globe familiar with the redeveloped intranet, he returned to academia to pursue his studies and passion for biologically-inspired computing.

Mr. Reiter is the recipient of two AMD Executive Spotlight awards for his modernisation of the AMD corporate intranet and deployment of a mobile employee engagement platform.